\documentclass[sigconf, 10pt]{acmart}

\AtBeginDocument{%
  \providecommand\BibTeX{{%
    \normalfont B\kern-0.5em{\scshape i\kern-0.25em b}\kern-0.8em\TeX}}}

\copyrightyear{2026}
\acmYear{2026}
\setcopyright{cc}
\setcctype{by-nc-nd}
\acmConference[GreenSys '26]{2nd International Workshop on Systems and Methods for Sustainable Large-Scale AI (GreenSys)}{April 27--30, 2026}{Edinburgh, Scotland UK}
\acmBooktitle{2nd International Workshop on Systems and Methods for Sustainable Large-Scale AI (GreenSys) (GreenSys '26), April 27--30, 2026, Edinburgh, Scotland UK}
\acmDOI{10.1145/3802973.3804458}
\acmISBN{979-8-4007-2174-8/2026/04}

\usepackage[utf8]{inputenc}
\usepackage[T1]{fontenc}
\usepackage{graphicx}
\usepackage{textcomp}

\usepackage{multirow}
\usepackage{rotating}

\begin{document}

\title{Energy Consumption Analysis of Discrete Diffusion and Autoregressive Language Models}

\author{Boris Ruf}
\affiliation{%
  \institution{AXA, AI Research}
  \city{Paris}
  \country{France}
}
\email{boris.ruf@axa.com}

\author{Marcin Detyniecki}
\affiliation{%
  \institution{AXA, AI Research}
  \city{Paris}
  \country{France}
}
\email{marcin.detyniecki@axa.com}

\begin{abstract}
As Large Language Models (LLMs) continue to scale, their energy consumption during inference has become a critical concern. While Autoregressive (AR) models are the industry standard, Diffusion Language Models (DLLMs) have emerged as a promising alternative capable of parallel token generation. In this paper, we perform a comparative energy benchmark of diffusion models (LLaDA-8B, Dream-7B) against AR counterparts (Llama-3.1-8B, Qwen-2.5-7B). We analyze the trade-offs between model architecture, KV-cache optimization, and Joules-per-token efficiency. Our results show that at longer sequences (200--600 tokens), diffusion models achieve 2.5--2.7 J/token, approximately 2$\times$ more efficient than AR models with KV-cache (5.3--5.7 J/token). KV-cache provides substantial benefits for AR models at longer sequences (50--70\% energy reduction).
\end{abstract}



\maketitle

\begin{figure}[t]
  \centering
  \includegraphics[width=\linewidth]{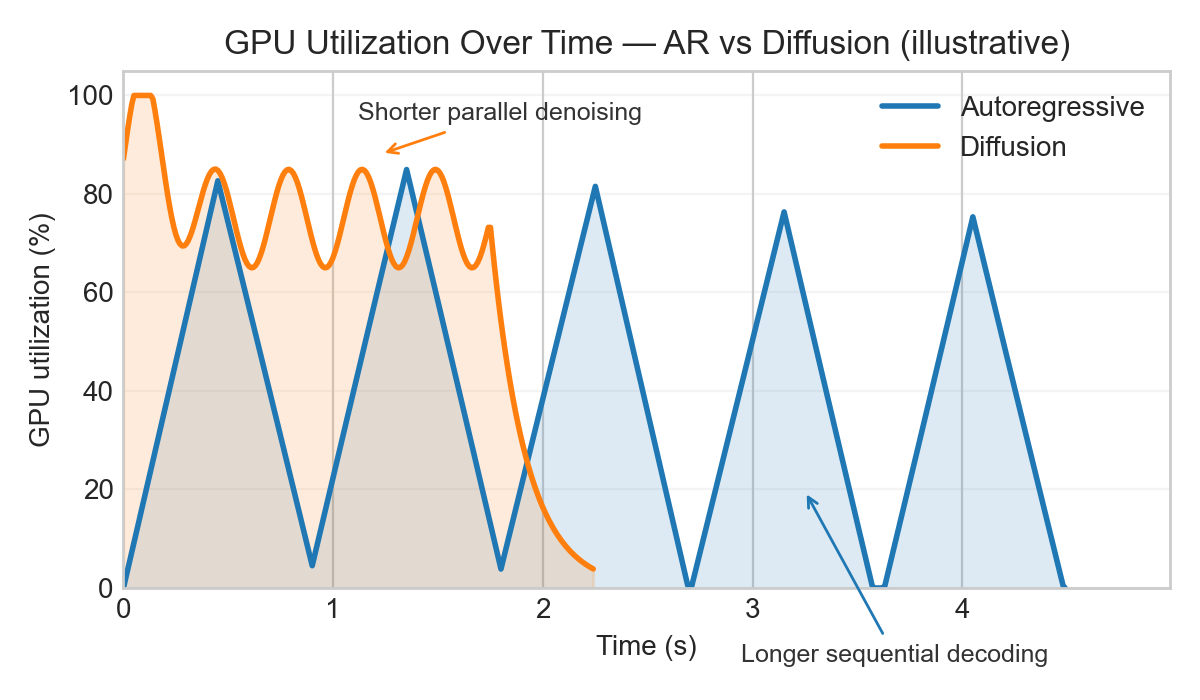}
  \caption{Illustrative model of GPU utilization over time for an autoregressive model (blue) with 10 output tokens vs a diffusion model (orange) with 10 denoising steps. This theoretical pattern motivates our empirical investigation.}
  \Description{Illustrative comparison of GPU utilization over time: the autoregressive model (blue) shows sequential peaks for each of 10 generated tokens, while the diffusion model (orange) shows a shorter, denser block corresponding to 10 denoising steps.}
  \label{fig:gpu-usage-illustrative}
\end{figure}

\section{Introduction}
Autoregressive language models generate text sequentially, which inherently limits parallelization and leads to high latency and energy costs for long sequences. Recently, text diffusion models have introduced the ability to refine multiple tokens simultaneously\cite{lin2026efficient}. However, the energy cost of multiple denoising steps versus a single-pass autoregressive generation remains under-explored.

Figure~\ref{fig:gpu-usage-illustrative} presents a theoretical model of GPU utilization for both architectures. The autoregressive pattern shows distinct, sequential peaks for each token, while the diffusion model exhibits a shorter, more intense block of parallel processing. This intuition suggests diffusion models may offer efficiency advantages---a hypothesis we investigate empirically in this paper.

We study the following primary research question:
\begin{quote}
  How does the end-to-end inference energy consumption (Joules-per-token) of discrete diffusion language models compare to autoregressive language models for instruction-style generation under matched experimental conditions?
\end{quote}

We operationalize this into three concrete sub-questions:
\begin{itemize}
  \item What is the Joules-per-token and latency-per-token of representative DLLMs (LLaDA, Dream) compared to AR baselines (Llama-3.1, Qwen-2.5) on instruction prompts?
  \item How does sequence length (10, 128, 1024 tokens) affect the relative energy efficiency of diffusion vs.\ autoregressive models?
  \item To what extent does KV-cache optimization reduce energy consumption for AR models, and does it close the efficiency gap with diffusion models?
\end{itemize}

The remainder of this paper is organized as follows: Section~\ref{sec:related} reviews related work on diffusion language models and energy measurement. Section~\ref{sec:setup} describes our experimental setup, including model selection, hardware configuration, and measurement protocol. Section~\ref{sec:results} presents our benchmark results. Section~\ref{sec:limitations} discusses limitations, Section~\ref{sec:future} outlines future work, followed by a broader discussion in Section~\ref{sec:discussion} and conclusions in Section~\ref{sec:conclusion}.

\section{Related Work}\label{sec:related}
Text diffusion for text generation has progressed from masked\slash discrete formulations such as MDLM \cite{sahoo2024mdlm} to large-scale diffusion LLMs and AR→dLLM conversion pipelines including DiffuLLaMA, Dream-7B, LLaDA, RND1, and Energy-Based Diffusion Language Models (EDLM) \cite{gong2025diffulama,ye2025dream7b,gsai2025llada,radical2025rnd1,xu2025edlm}. Work on AR→block-diffusion and light-weight adaptation (e.g., SDAR, Fast-dLLM) demonstrates that structural alignment with AR pretraining can substantially reduce adaptation cost compared to training diffusion models from scratch \cite{cheng2025sdar,wu2026fastdllm,arriola2025block}.

At inference time, dLLMs enable parallel token updates but introduce system-level bottlenecks: bidirectional attention patterns complicate KV-cache reuse, and many samplers require multiple denoising steps, increasing end-to-end compute. Research addressing these challenges includes parallel decoding and sampler-distillation approaches (e.g., dParallel~\cite{chen2025dparallel}), adaptive KV strategies (dKV-Cache~\cite{ma2025dkvcache}), and block/anchor decoding designs aimed at reducing sampling steps while preserving quality~\cite{wu2026fastdllm}.

There is also a growing literature on measuring and reporting energy and carbon for large-scale NLP. Seminal studies established that large training and evaluation runs can yield substantial energy and emissions, motivating the "Green AI" advocacy for efficiency-aware reporting \cite{strubell2019energy,schwartz2019green}. Practical tooling such as CodeCarbon enables reproducible runtime energy logging at the hardware level \cite{codecarbon2021}. Very recent studies provide hardware-level profiling for diffusion-based models in the image domain and show substantial variance across samplers, denoising schedules, and backends \cite{bertazzini2025hiddencostimagequantifying,iyengar2025energyscalinglawsdiffusion}.

Despite these advances, hardware-level Joules-per-token measurements for text dLLMs are still scarce. This paper fills that gap by reporting end-to-end energy (Joules-per-token) and latency under controlled cache and backend settings, and by situating the measured trade-offs alongside the algorithmic and systems techniques described above.

\section{Experimental Setup}\label{sec:setup}
This section details the hardware, software, and measurement protocol used in our study so that results are reproducible.

\subsection{Model selection}
We evaluate a compact set of diffusion and autoregressive models chosen to illustrate different algorithmic and system trade-offs:
\begin{itemize}
  \item \textbf{LLaDA-8B} --- a diffusion language model trained from scratch using a Llama-style architecture; chosen as a representative dLLM that demonstrates native diffusion training can achieve performance competitive with AR models \cite{gsai2025llada}.
  \item \textbf{Dream-7B} --- a diffusion LLM designed for efficient denoising schedules and parallel sampling; used to evaluate step-count vs. energy trade-offs \cite{ye2025dream7b}.
  \item \textbf{Llama-3.1-8B-Instruct} --- a modern autoregressive foundation model serving as a large AR baseline for instruction-style generation \cite{touvron2023llama}.
  \item \textbf{Qwen-2.5-7B-Instruct} --- an autoregressive instruction-tuned baseline representative of high-quality open-source models in the same parameter regime as some diffusion models \cite{qwen2024model}.
\end{itemize}

\subsection{Hardware and system configuration}
All inference runs reported in this paper were executed on a Microsoft Azure compute instance equipped with an NVIDIA A100. We use mixed precision (bf16/float16) where supported and ensure consistent CUDA/cuDNN and PyTorch versions across runs. We run models inside per-model Docker images to lock runtime dependencies (notably different `transformers` versions). We measure energy by querying NVIDIA's NVML API to capture GPU power draw and integrate over the measured duration to obtain kWh/Joules. 

\begin{table*}[t]
  \centering
  \caption{Energy and latency per generated token across different sequence lengths. AR models shown with and without KV-cache. Diffusion models do not use KV-cache. Note: Qwen-2.5 generated fewer tokens at max=1024 because the model produced an end-of-sequence token before reaching the limit.}
  \label{tab:results-summary}
  \small
  \begin{tabular}{c l c r r r r}
    \toprule
    \makebox[0em]{} & Model & KV-Cache & Max Tokens & Gen. Tokens & J/token & s/token \\
    \midrule
    \multicolumn{7}{l}{\textit{Short sequences (max\_tokens = 10)}} \\
    \multirow{4}{*}{\rotatebox{90}{AR}} 
    & Llama-3.1-8B-Instruct & Off & 10 & 10 & 8.32 & 0.0300 \\
    & Llama-3.1-8B-Instruct & On & 10 & 10 & 8.10 & 0.0295 \\
    & Qwen-2.5-7B-Instruct & Off & 10 & 10 & 7.93 & 0.0270 \\
    & Qwen-2.5-7B-Instruct & On & 10 & 10 & 7.92 & 0.0265 \\
    \addlinespace
    \multirow{2}{*}{\rotatebox{90}{Diff}} 
    & Dream-7B & --- & 10 & 10 & 32.21 & 0.1676 \\
    & LLaDA-8B & --- & 10 & 10 & 35.60 & 0.1767 \\
    \midrule
    \multicolumn{7}{l}{\textit{Medium sequences (max\_tokens = 128)}} \\
    \multirow{4}{*}{\rotatebox{90}{AR}} 
    & Llama-3.1-8B-Instruct & Off & 128 & 128 & 6.73 & 0.0294 \\
    & Llama-3.1-8B-Instruct & On & 128 & 128 & 5.07 & 0.0286 \\
    & Qwen-2.5-7B-Instruct & Off & 128 & 128 & 6.63 & 0.0261 \\
    & Qwen-2.5-7B-Instruct & On & 128 & 128 & 5.46 & 0.0266 \\
    \addlinespace
    \multirow{2}{*}{\rotatebox{90}{Diff}} 
    & Dream-7B & --- & 128 & 128 & 4.45 & 0.0136 \\
    & LLaDA-8B & --- & 128 & 128 & 5.11 & 0.0156 \\
    \midrule
    \multicolumn{7}{l}{\textit{Long sequences (max\_tokens = 1024)}} \\
    \multirow{4}{*}{\rotatebox{90}{AR}} 
    & Llama-3.1-8B-Instruct & Off & 1024 & 1024 & 18.21 & 0.0600 \\
    & Llama-3.1-8B-Instruct & On & 1024 & 1024 & 5.26 & 0.0272 \\
    & Qwen-2.5-7B-Instruct & Off & 1024 & 564$^\dagger$ & 12.51 & 0.0373 \\
    & Qwen-2.5-7B-Instruct & On & 1024 & 587$^\dagger$ & 5.72 & 0.0267 \\
    \addlinespace
    \multirow{2}{*}{\rotatebox{90}{Diff}} 
    & Dream-7B & --- & 1024 & 1024 & 2.01 & 0.0064 \\
    & LLaDA-8B & --- & 1024 & 1024 & 2.40 & 0.0068 \\
    \bottomrule
    \multicolumn{7}{l}{\footnotesize $^\dagger$Model generated EOS token before reaching max\_tokens limit.}
  \end{tabular}
\end{table*}

\subsection{Evaluation protocol}
We use a short, fixed instruction prompt for all runs. Each model configuration was evaluated at three target sequence lengths: 10, 128, and 1024 tokens. For each configuration, we perform 5 warmup runs followed by 10 measured runs. AR models were tested both with and without KV-cache enabled.

\subsection{Metrics}
We report the following primary metrics:
\begin{itemize}
  \item \textbf{Energy per token (J/token):} GPU power draw captured via NVIDIA's NVML API, integrated over run duration, converted from kWh to Joules ($1\ \mathrm{kWh}=3.6\times10^{6}\ \mathrm{J}$), and divided by generated tokens.
  \item \textbf{Latency per token (s/token):} Wall-clock time divided by generated tokens.
\end{itemize}
Results are reported as means over the 10 measured runs. Energy measurements include total GPU power draw during generation; no idle baseline was subtracted.

\section{Results}\label{sec:results}
Table~\ref{tab:results-summary} presents energy and latency per generated token across three sequence lengths (10, 128, and 1024 tokens), revealing a critical crossover point: AR models are more efficient at short sequences, but diffusion models dominate at longer sequences.

\subsection{Per-model observations}
Key observations derived from the measured means and supporting figures are:
\begin{itemize}
  \item \textbf{Autoregressive models (Llama-3.1, Qwen-2.5)}: AR models show relatively stable per-token efficiency across sequence lengths when KV-cache is enabled ($\approx$5--8 J/token). Without cache, efficiency degrades substantially at longer sequences (18.21 J/token for Llama at 1024 tokens). Qwen-2.5 autonomously terminated generation early (564--587 tokens) by producing an end-of-sequence token, demonstrating that max\_tokens is an upper bound, not a guarantee.
  \item \textbf{Diffusion models (Dream-7B, LLaDA-8B)}: Efficiency improves dramatically with sequence length due to amortization of fixed initialization costs. At 10 tokens, diffusion models are inefficient (32--36 J/token), but at 1024 tokens they achieve 2.01--2.40 J/token---the most efficient results in our benchmark.
\end{itemize}

\begin{figure*}[t]
  \centering
  \includegraphics[width=\linewidth,height=0.35\textheight,keepaspectratio]{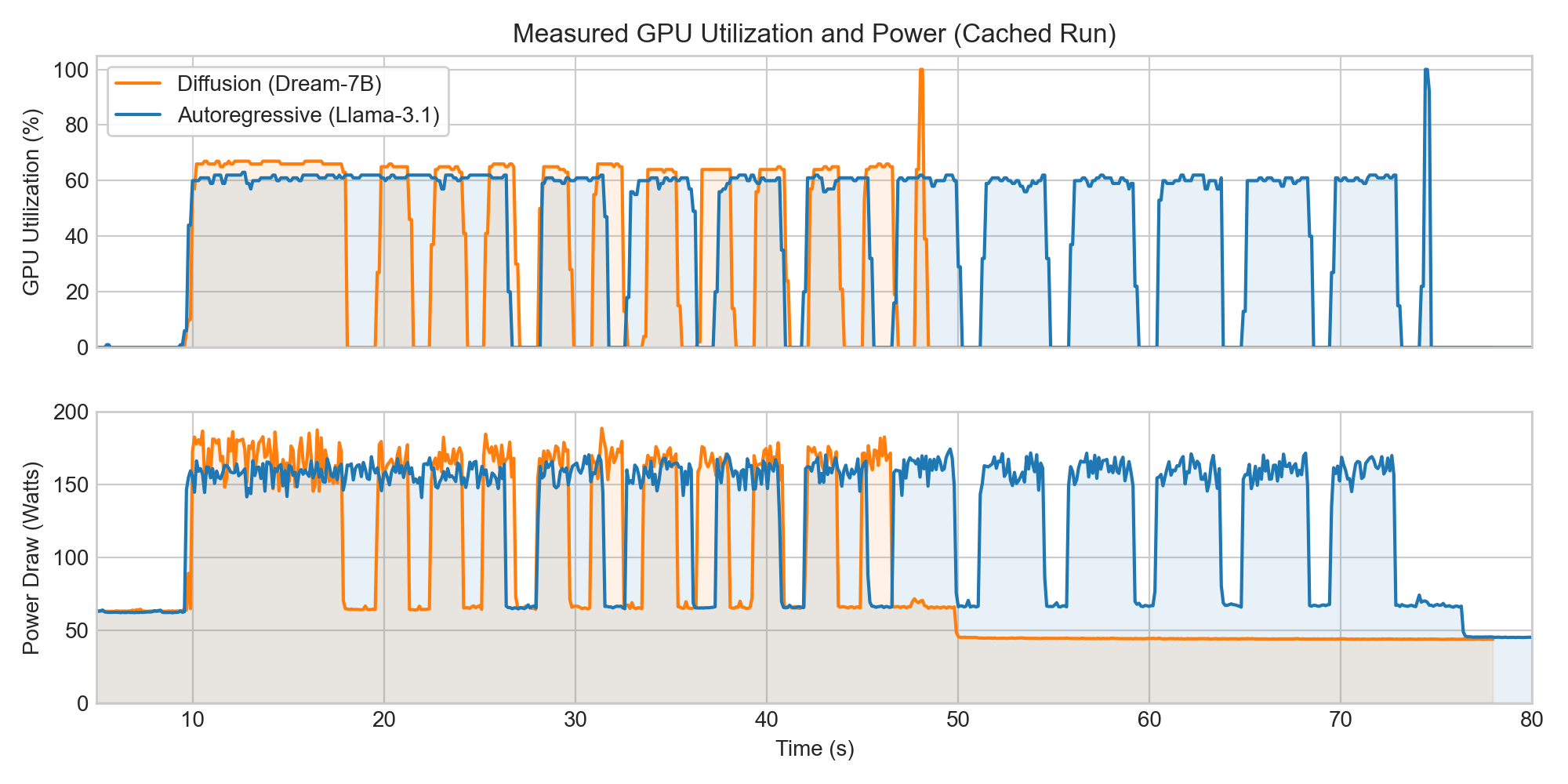}
  \caption{Measured GPU utilization and power draw for a cached run, comparing an autoregressive model (Llama-3.1) with a diffusion model (Dream-7B). This plot visualizes the execution from our traces where the autoregressive model's KV-cache is active. The initial period ($\approx$8s for diffusion, $\approx$18s for AR) captures the benchmark's warmup phase. The subsequent, stable utilization pattern corresponds to the 10 measured runs, during which each model generated 128 tokens per run.}
  \Description{Measured GPU utilization and power traces for cached runs comparing Llama-3.1 (autoregressive) and Dream-7B (diffusion). The first segment is warmup (about 8 seconds for diffusion and 18 seconds for autoregressive), followed by a stable repeated pattern from 10 measured runs of 128-token generation per run.}
  \label{fig:gpu-usage-cached}
\end{figure*}

\subsection{Impact of KV-cache on AR models}
Autoregressive models benefit substantially from KV-cache, which stores previously computed key-value pairs to avoid redundant computation during sequential decoding.

At short sequences (10 tokens), KV-cache provides minimal benefit: Llama drops from 8.32 to 8.10 J/token (3\% reduction). At long sequences (1024 tokens), the benefit is dramatic: Llama drops from 18.21 to 5.26 J/token (71\% reduction), and Qwen from 12.51 to 5.72 J/token (54\% reduction).

Diffusion models cannot exploit KV-cache due to their bidirectional attention patterns. However, their efficiency at long sequences (2.01--2.40 J/token) still exceeds cached AR models (5.26--5.72 J/token).

\subsection{Diffusion vs. Autoregressive}\label{sec:diffusion-vs-ar}
The relative efficiency of diffusion vs.\ AR models depends critically on sequence length:

\textbf{Short sequences (10 tokens):} AR models are significantly more efficient. With cache, Llama achieves 8.10 J/token vs.\ 32.21 J/token for Dream-7B---a 4$\times$ advantage for AR.

\textbf{Medium sequences (128 tokens):} Crossover occurs. Dream-7B (4.45 J/token) outperforms cached Llama (5.07 J/token), while LLaDA (5.11 J/token) is comparable.

\textbf{Long sequences (1024 tokens):} Diffusion models dominate. Dream-7B achieves 2.01 J/token and LLaDA achieves 2.40 J/token, compared to 5.26--5.72 J/token for cached AR models---approximately 2.5$\times$ more efficient. Without cache, AR models degrade to 12--18 J/token.

Latency follows the same pattern: at 1024 tokens, diffusion models achieve 0.006--0.007 s/token vs.\ 0.027 s/token for cached AR---a 4$\times$ advantage due to parallel token generation.

Figure~\ref{fig:gpu-usage-cached} shows measured GPU utilization and power draw, illustrating the diffusion model's shorter, more intense compute pattern compared to the AR model's sequential token generation.

\section{Limitations}\label{sec:limitations}
Our study uses a single fixed prompt and three sequence lengths (10, 128, 1024 tokens), which identifies a crossover around 128 tokens but leaves the precise threshold uncertain. A finer-grained sweep would pinpoint efficiency ``cliffs'' more precisely. We also did not vary diffusion hyperparameters: investigating how denoising step counts (e.g., 8--128 steps) impact J/token would clarify quality-efficiency trade-offs. Additionally, while published benchmarks suggest that the diffusion models evaluated here (Dream-7B, LLaDA-8B) achieve accuracy comparable to their AR counterparts, this study did not rigorously evaluate output quality; future work should incorporate benchmarks (MMLU, HellaSwag) to verify that efficiency gains do not come at the cost of generation quality. Finally, all experiments ran on a single A100 using vanilla HuggingFace Transformers---results may differ on other hardware or with optimized backends (vLLM, TensorRT-LLM).

\section{Future Work}\label{sec:future}
Beyond these limitations, we see a concrete systems direction: energy-aware adaptive inference. A next step is to map the crossover point more precisely with finer output-length sweeps and to evaluate quality--energy trade-offs jointly, rather than energy alone. At the systems level, our results motivate a hybrid serving policy that routes requests to AR or diffusion backends based on expected output length, latency constraints, and current load. Such a policy should be validated across hardware platforms, optimized runtimes, and quantization settings to test robustness in production-like conditions.

\section{Discussion}\label{sec:discussion}
The crossover pattern observed in Section~\ref{sec:results} arises from the different computational characteristics of each architecture. Diffusion models incur fixed initialization overhead that is amortized over longer sequences, while their parallel generation becomes advantageous. AR models, even with KV-cache, maintain relatively constant per-token costs that become less competitive at scale.

Note that Qwen-2.5 generated fewer tokens (564--587) at max\_tokens=1024 because the model produced an EOS token before reaching the limit. This is expected behavior: max\_tokens specifies an upper bound, and models may naturally conclude responses earlier. However, this complicates direct energy comparisons, as the workload differs.

Our KV-cache experiments confirm that cache provides substantial benefits for AR models (50--70\% energy reduction at long sequences), but even with this optimization, diffusion models maintain their efficiency advantage at longer sequences. In practice, these findings support adaptive architecture selection rather than universally preferring a single model family.

\section{Conclusion}\label{sec:conclusion}
This study provides an energy comparison between diffusion and autoregressive language models across different sequence lengths. Our key finding is a crossover in efficiency:

\begin{itemize}
  \item \textbf{Short sequences ($\leq$10 tokens):} AR models are 4$\times$ more efficient (8 J/token vs.\ 32--36 J/token)
  \item \textbf{Medium sequences ($\approx$128 tokens):} Efficiency converges; diffusion begins to lead
  \item \textbf{Long sequences ($\geq$1024 tokens):} Diffusion models are 2.5$\times$ more efficient (2 J/token vs.\ 5 J/token)
\end{itemize}

At long sequences, diffusion models also achieve 4$\times$ lower latency (0.006 s/token vs.\ 0.027 s/token) due to parallel generation. KV-cache provides substantial benefits for AR models but does not close the gap at longer sequences.

These results suggest that deployment choices should consider expected output length: AR models remain preferable for short responses, while diffusion models offer compelling advantages for longer generation tasks.

\bibliographystyle{ACM-Reference-Format}
\bibliography{bibliography}


\begin{thebibliography}{19}


\ifx \showCODEN    \undefined \def \showCODEN     #1{\unskip}     \fi
\ifx \showISBNx    \undefined \def \showISBNx     #1{\unskip}     \fi
\ifx \showISBNxiii \undefined \def \showISBNxiii  #1{\unskip}     \fi
\ifx \showISSN     \undefined \def \showISSN      #1{\unskip}     \fi
\ifx \showLCCN     \undefined \def \showLCCN      #1{\unskip}     \fi
\ifx \shownote     \undefined \def \shownote      #1{#1}          \fi
\ifx \showarticletitle \undefined \def \showarticletitle #1{#1}   \fi
\ifx \showURL      \undefined \def \showURL       {\relax}        \fi
\providecommand\bibfield[2]{#2}
\providecommand\bibinfo[2]{#2}
\providecommand\natexlab[1]{#1}
\providecommand\showeprint[2][]{arXiv:#2}

\bibitem[Arriola et~al\mbox{.}(2025)]%
        {arriola2025block}
\bibfield{author}{\bibinfo{person}{Marianne Arriola}, \bibinfo{person}{Aaron
  Gokaslan}, \bibinfo{person}{Justin~T Chiu}, \bibinfo{person}{Jiaqi Han},
  \bibinfo{person}{Zhihan Yang}, \bibinfo{person}{Zhixuan Qi},
  \bibinfo{person}{Subham~Sekhar Sahoo}, {and} \bibinfo{person}{Volodymyr
  Kuleshov}.} \bibinfo{year}{2025}\natexlab{}.
\newblock \showarticletitle{Interpolating Autoregressive and Discrete Denoising
  Diffusion Language Models}. In \bibinfo{booktitle}{\emph{ICLR}}.
\newblock


\bibitem[Bertazzini et~al\mbox{.}(2025)]%
        {bertazzini2025hiddencostimagequantifying}
\bibfield{author}{\bibinfo{person}{Giulia Bertazzini}, \bibinfo{person}{Chiara
  Albisani}, \bibinfo{person}{Daniele Baracchi}, \bibinfo{person}{Dasara
  Shullani}, {and} \bibinfo{person}{Roberto Verdecchia}.}
  \bibinfo{year}{2025}\natexlab{}.
\newblock \bibinfo{title}{The Hidden Cost of an Image: Quantifying the Energy
  Consumption of AI Image Generation}.
\newblock


\bibitem[Chandrasegaran et~al\mbox{.}(2025)]%
        {radical2025rnd1}
\bibfield{author}{\bibinfo{person}{Keshigeyan Chandrasegaran},
  \bibinfo{person}{Armin~W. Thomas}, \bibinfo{person}{Jerome Ku},
  \bibinfo{person}{Federico Berto}, \bibinfo{person}{Jae~Myung Kim},
  \bibinfo{person}{Garyk Brixi}, \bibinfo{person}{Eric Nguyen},
  \bibinfo{person}{Stefano Massaroli}, {and} \bibinfo{person}{Michael Poli}.}
  \bibinfo{year}{2025}\natexlab{}.
\newblock \bibinfo{title}{RND1: Simple, Scalable AR-to-Diffusion Conversion}.
\newblock


\bibitem[Chen et~al\mbox{.}(2026)]%
        {chen2025dparallel}
\bibfield{author}{\bibinfo{person}{Zigeng Chen}, \bibinfo{person}{Gongfan
  Fang}, \bibinfo{person}{Xinyin Ma}, \bibinfo{person}{Ruonan Yu}, {and}
  \bibinfo{person}{Xinchao Wang}.} \bibinfo{year}{2026}\natexlab{}.
\newblock \showarticletitle{dParallel: Learnable Parallel Decoding for dLLMs}.
  In \bibinfo{booktitle}{\emph{ICLR}}.
\newblock


\bibitem[Cheng et~al\mbox{.}(2025)]%
        {cheng2025sdar}
\bibfield{author}{\bibinfo{person}{Shuang Cheng}, \bibinfo{person}{Yihan Bian},
  \bibinfo{person}{Dawei Liu}, \bibinfo{person}{Linfeng Zhang},
  \bibinfo{person}{Qian Yao}, \bibinfo{person}{Zhongbo Tian},
  \bibinfo{person}{Wenhai Wang}, \bibinfo{person}{Qipeng Guo},
  \bibinfo{person}{Kai Chen}, \bibinfo{person}{Biqing Qi}, {and}
  \bibinfo{person}{Bowen Zhou}.} \bibinfo{year}{2025}\natexlab{}.
\newblock \showarticletitle{SDAR: A Synergistic Diffusion-AutoRegression
  Paradigm for Scalable Sequence Generation}.
\newblock \bibinfo{journal}{\emph{CoRR}} (\bibinfo{year}{2025}).
\newblock


\bibitem[Courty et~al\mbox{.}(2024)]%
        {codecarbon2021}
\bibfield{author}{\bibinfo{person}{Benoit Courty}, \bibinfo{person}{Victor
  Schmidt}, \bibinfo{person}{Sasha Luccioni}, \bibinfo{person}{Goyal-Kamal},
  \bibinfo{person}{Marion Coutarel}, \bibinfo{person}{Boris Feld},
  \bibinfo{person}{J{\'e}r{\'e}my Lecourt}, \bibinfo{person}{Liam Connell},
  \bibinfo{person}{Amine Saboni}, {et~al\mbox{.}}}
  \bibinfo{year}{2024}\natexlab{}.
\newblock \bibinfo{title}{mlco2/codecarbon: v2.4.1}.
\newblock


\bibitem[Gong et~al\mbox{.}(2025)]%
        {gong2025diffulama}
\bibfield{author}{\bibinfo{person}{Shansan Gong}, \bibinfo{person}{Shivam
  Agarwal}, \bibinfo{person}{Yizhe Zhang}, \bibinfo{person}{Jiacheng Ye},
  \bibinfo{person}{Lin Zheng}, \bibinfo{person}{Mukai Li},
  \bibinfo{person}{Chenxin An}, \bibinfo{person}{Peilin Zhao},
  \bibinfo{person}{Wei Bi}, \bibinfo{person}{Jiawei Han}, \bibinfo{person}{Hao
  Peng}, {and} \bibinfo{person}{Lingpeng Kong}.}
  \bibinfo{year}{2025}\natexlab{}.
\newblock \showarticletitle{Scaling Diffusion Language Models via Adaptation
  from Autoregressive Models}. In \bibinfo{booktitle}{\emph{ICLR}}.
\newblock


\bibitem[Iyengar et~al\mbox{.}(2025)]%
        {iyengar2025energyscalinglawsdiffusion}
\bibfield{author}{\bibinfo{person}{Aniketh Iyengar}, \bibinfo{person}{Jiaqi
  Han}, \bibinfo{person}{Boris Ruf}, \bibinfo{person}{Vincent Grari},
  \bibinfo{person}{Marcin Detyniecki}, {and} \bibinfo{person}{Stefano Ermon}.}
  \bibinfo{year}{2025}\natexlab{}.
\newblock \bibinfo{title}{Energy Scaling Laws for Diffusion Models: Quantifying
  Compute and Carbon Emissions in Image Generation}.
\newblock


\bibitem[Lin et~al\mbox{.}(2026)]%
        {lin2026efficient}
\bibfield{author}{\bibinfo{person}{Haokun Lin}, \bibinfo{person}{Xinle Jia},
  \bibinfo{person}{Shaozhen Liu}, \bibinfo{person}{Shujun Xia},
  \bibinfo{person}{Weitao Huang}, \bibinfo{person}{Haobo Xu},
  \bibinfo{person}{Junyang Li}, \bibinfo{person}{Yicheng Xiao},
  \bibinfo{person}{Xingrun Xing}, \bibinfo{person}{Ziyu Guo}, {et~al\mbox{.}}}
  \bibinfo{year}{2026}\natexlab{}.
\newblock \showarticletitle{Efficient Diffusion Language Models: A
  Comprehensive Survey}.
\newblock \bibinfo{journal}{\emph{Authorea Preprints}} (\bibinfo{year}{2026}).
\newblock


\bibitem[Ma et~al\mbox{.}(2025)]%
        {ma2025dkvcache}
\bibfield{author}{\bibinfo{person}{Xinyin Ma}, \bibinfo{person}{Runpeng Yu},
  \bibinfo{person}{Gongfan Fang}, {and} \bibinfo{person}{Xinchao Wang}.}
  \bibinfo{year}{2025}\natexlab{}.
\newblock \showarticletitle{d{KV}-Cache: The Cache for Diffusion Language
  Models}. In \bibinfo{booktitle}{\emph{NeurIPS}}.
\newblock


\bibitem[Nie et~al\mbox{.}(2025)]%
        {gsai2025llada}
\bibfield{author}{\bibinfo{person}{Shen Nie}, \bibinfo{person}{Fengqi Zhu},
  \bibinfo{person}{Zebin You}, \bibinfo{person}{Xiaolu Zhang},
  \bibinfo{person}{Jingyang Ou}, \bibinfo{person}{Jun Hu}, \bibinfo{person}{Jun
  Zhou}, \bibinfo{person}{Yankai Lin}, \bibinfo{person}{Ji-Rong Wen}, {and}
  \bibinfo{person}{Chongxuan Li}.} \bibinfo{year}{2025}\natexlab{}.
\newblock \showarticletitle{Large language diffusion models}.
\newblock \bibinfo{journal}{\emph{arXiv}} (\bibinfo{year}{2025}).
\newblock


\bibitem[Sahoo et~al\mbox{.}(2024)]%
        {sahoo2024mdlm}
\bibfield{author}{\bibinfo{person}{Subham~Sekhar Sahoo},
  \bibinfo{person}{Marianne Arriola}, \bibinfo{person}{Yair Schiff},
  \bibinfo{person}{Aaron Gokaslan}, \bibinfo{person}{Edgar Marroquin},
  \bibinfo{person}{Justin~T Chiu}, \bibinfo{person}{Alexander Rush}, {and}
  \bibinfo{person}{Volodymyr Kuleshov}.} \bibinfo{year}{2024}\natexlab{}.
\newblock \showarticletitle{Simple and effective masked diffusion language
  models}. In \bibinfo{booktitle}{\emph{NeurIPS}}.
\newblock


\bibitem[Schwartz et~al\mbox{.}(2020)]%
        {schwartz2019green}
\bibfield{author}{\bibinfo{person}{Roy Schwartz}, \bibinfo{person}{Jesse
  Dodge}, \bibinfo{person}{Noah~A. Smith}, {and} \bibinfo{person}{Oren
  Etzioni}.} \bibinfo{year}{2020}\natexlab{}.
\newblock \showarticletitle{Green AI}.
\newblock \bibinfo{journal}{\emph{Commun. ACM}} (\bibinfo{year}{2020}).
\newblock


\bibitem[Strubell et~al\mbox{.}(2019)]%
        {strubell2019energy}
\bibfield{author}{\bibinfo{person}{Emma Strubell}, \bibinfo{person}{Ananya
  Ganesh}, {and} \bibinfo{person}{Andrew McCallum}.}
  \bibinfo{year}{2019}\natexlab{}.
\newblock \showarticletitle{Energy and policy considerations for deep learning
  in NLP}. In \bibinfo{booktitle}{\emph{ACL}}.
\newblock


\bibitem[Touvron et~al\mbox{.}(2023)]%
        {touvron2023llama}
\bibfield{author}{\bibinfo{person}{Hugo Touvron}, \bibinfo{person}{Thibaut
  Lavril}, \bibinfo{person}{Gautier Izacard}, \bibinfo{person}{Xavier
  Martinet}, \bibinfo{person}{Marie-Anne Lachaux},
  \bibinfo{person}{Timoth{\'e}e Lacroix}, \bibinfo{person}{Baptiste
  Rozi{\`e}re}, \bibinfo{person}{Naman Goyal}, \bibinfo{person}{Eric Hambro},
  \bibinfo{person}{Faisal Azhar}, {et~al\mbox{.}}}
  \bibinfo{year}{2023}\natexlab{}.
\newblock \showarticletitle{LLaMA: Open and Efficient Foundation Language
  Models}.
\newblock \bibinfo{journal}{\emph{arXiv}} (\bibinfo{year}{2023}).
\newblock


\bibitem[Wu et~al\mbox{.}(2025)]%
        {wu2026fastdllm}
\bibfield{author}{\bibinfo{person}{Chengyue Wu}, \bibinfo{person}{Hao Zhang},
  \bibinfo{person}{Shuchen Xue}, \bibinfo{person}{Shizhe Diao},
  \bibinfo{person}{Yonggan Fu}, \bibinfo{person}{Zhijian Liu},
  \bibinfo{person}{Pavlo Molchanov}, \bibinfo{person}{Ping Luo},
  \bibinfo{person}{Song Han}, {and} \bibinfo{person}{Enze Xie}.}
  \bibinfo{year}{2025}\natexlab{}.
\newblock \bibinfo{title}{Fast-dLLM v2: Efficient Block-Diffusion LLM}.
\newblock


\bibitem[Xu et~al\mbox{.}(2025)]%
        {xu2025edlm}
\bibfield{author}{\bibinfo{person}{Minkai Xu}, \bibinfo{person}{Tomas Geffner},
  \bibinfo{person}{Karsten Kreis}, \bibinfo{person}{Weili Nie},
  \bibinfo{person}{Yilun Xu}, \bibinfo{person}{Jure Leskovec},
  \bibinfo{person}{Stefano Ermon}, {and} \bibinfo{person}{Arash Vahdat}.}
  \bibinfo{year}{2025}\natexlab{}.
\newblock \showarticletitle{Energy-based Diffusion Language Models for Text
  Generation}. In \bibinfo{booktitle}{\emph{ICLR}}.
\newblock


\bibitem[Yang et~al\mbox{.}(2024)]%
        {qwen2024model}
\bibfield{author}{\bibinfo{person}{An Yang}, \bibinfo{person}{Baosong Yang},
  \bibinfo{person}{Beichen Zhang}, \bibinfo{person}{Binyuan Hui},
  \bibinfo{person}{Bo Zheng}, \bibinfo{person}{Bowen Yu},
  \bibinfo{person}{Chengyuan Li}, \bibinfo{person}{Dayiheng Liu},
  \bibinfo{person}{Fei Huang}, \bibinfo{person}{Haoran Wei},
  \bibinfo{person}{Huan Lin}, \bibinfo{person}{Jian Yang},
  \bibinfo{person}{Jianhong Tu}, \bibinfo{person}{Jianwei Zhang},
  \bibinfo{person}{Jianxin Yang}, \bibinfo{person}{Jiaxi Yang},
  \bibinfo{person}{Jingren Zhou}, \bibinfo{person}{Junyang Lin},
  \bibinfo{person}{Kai Dang}, \bibinfo{person}{Keming Lu},
  \bibinfo{person}{Keqin Bao}, \bibinfo{person}{Kexin Yang},
  \bibinfo{person}{Le Yu}, \bibinfo{person}{Mei Li}, \bibinfo{person}{Mingfeng
  Xue}, \bibinfo{person}{Pei Zhang}, \bibinfo{person}{Qin Zhu},
  \bibinfo{person}{Rui Men}, \bibinfo{person}{Runji Lin},
  \bibinfo{person}{Tianhao Li}, \bibinfo{person}{Tingyu Xia},
  \bibinfo{person}{Xingzhang Ren}, \bibinfo{person}{Xuancheng Ren},
  \bibinfo{person}{Yang Fan}, \bibinfo{person}{Yang Su},
  \bibinfo{person}{Yichang Zhang}, \bibinfo{person}{Yu Wan},
  \bibinfo{person}{Yuqiong Liu}, \bibinfo{person}{Zeyu Cui},
  \bibinfo{person}{Zhenru Zhang}, {and} \bibinfo{person}{Zihan Qiu}.}
  \bibinfo{year}{2024}\natexlab{}.
\newblock \showarticletitle{Qwen2.5 Technical Report}.
\newblock \bibinfo{journal}{\emph{CoRR}} (\bibinfo{year}{2024}).
\newblock


\bibitem[Ye et~al\mbox{.}(2025)]%
        {ye2025dream7b}
\bibfield{author}{\bibinfo{person}{Jiacheng Ye}, \bibinfo{person}{Zhihui Xie},
  \bibinfo{person}{Lin Zheng}, \bibinfo{person}{Jiahui Gao},
  \bibinfo{person}{Zirui Wu}, \bibinfo{person}{Xin Jiang},
  \bibinfo{person}{Zhenguo Li}, {and} \bibinfo{person}{Lingpeng Kong}.}
  \bibinfo{year}{2025}\natexlab{}.
\newblock \showarticletitle{Dream 7B: Diffusion Large Language Models}.
\newblock \bibinfo{journal}{\emph{arXiv}} (\bibinfo{year}{2025}).
\newblock


\end{thebibliography}

\end{document}